\documentclass{article}
\usepackage{icml2009} 
\usepackage{graphicx} % more modern
\usepackage{subfigure} 
\usepackage{mlapa}
\usepackage{amsmath, amsthm, amssymb}

\icmltitlerunning{Robust Boosting}

\newcommand{\newmcommand}[2]{\newcommand{#1}{{\ifmmode {#2}\else\mbox{${#2}$}\fi}}}
\newcommand{\renewmcommand}[2]{\renewcommand{#1}{{\ifmmode {#2}\else\mbox{${#2}$}\fi}}}
\newcommand{\newmcommandi}[2]{\newcommand{#1}[1]{{\ifmmode {#2}\else\mbox{${#2}$}\fi}}}
\newcommand{\newmcommandii}[2]{\newcommand{#1}[2]{{\ifmmode {#2}\else\mbox{${#2}$}\fi}}}
\newcommand{\newmcommandiii}[2]{\newcommand{#1}[3]{{\ifmmode {#2}\else\mbox{${#2}$}\fi}}}

\newfont{\msym}{msbm10}

\newcommand{\paren}[1]{\left({#1}\right)}
\newcommand{\brackets}[1]{\left[{#1}\right]}

\newcommand{\sign}{{\rm sign}}

\newcommand{\IndF}[1]{{\bf 1}\brackets{#1}} % Indicator Function

\newcommand{\X}{{\cal X}}      % input space
\newcommand{\Exp}[2]{{\rm E}_{#1}\brackets{#2}} % Expected value of RV
\newcommand{\hf}{H}  % final classifier

\newcommand{\hyp}{{\cal H}} % space of base classifiers

\newmcommand{\M}{\bf M}
\newmcommand{\dM}{\M'}
\newmcommand{\D}{{\cal D}}
\renewmcommand{\P}{P}
\newmcommand{\Q}{Q}
\newmcommand{\Pt}{\P_t}
\newmcommand{\Qt}{\Q_t}
\newmcommand{\Pstar}{\P^*}
\newmcommand{\Pref}{\tilde{\P}}	% a reference mixed strategy (not
\newmcommand{\Qstar}{\Q^*}
\newmcommand{\Pa}{\overline{\P}}
\newmcommand{\Qa}{\overline{\Q}}
\newmcommandi{\trans}{{#1}^{\top}}
\newmcommand{\mhx}{\M(h,x)}
\newmcommand{\mxh}{\dM(x,h)}
\newmcommand{\mpq}{\M(\P,\Q)}
\newmcommand{\mpsq}{\M(\Pstar,\Q)}
\newmcommand{\mpsqt}{\M(\Pstar,\Qt)}
\newmcommand{\mptqt}{\M(\Pt,\Qt)}
\newmcommand{\mptq}{\M(\Pt,\Q)}
\newmcommand{\mpqt}{\M(\P,\Qt)}

\newmcommand{\sumt}{\sum_{t=1}^T}
\newmcommand{\sumim}{\sum_{i=1}^m}
\newmcommand{\delt}{\Delta_T}

\newmcommand{\hyps}{\hyp}
\newmcommand{\predt}{\hat{y}_t}
\newmcommandii{\prob}{\Pr_{#1}\brackets{{#2}}}

\newmcommand{\hfin}{\hf}

\newcommand{\N}{N}		% number of training examples 

\newcommand{\rpot}[2]{\Phi(#1,#2)}
\newcommand{\rw}[2]{w(#1,#2)}
\newcommand{\deltat}{\Delta t}
\newcommand{\s}{m}
\newcommand{\dels}{\Delta m}

\newcommand{\erf}{\mbox{erf}}

\newcommand{\sigmaf}{\sigma_f}

\begin{document} 

\twocolumn[
\icmltitle{A more robust boosting algorithm}

\icmlauthor{Yoav Freund}{yfreund@ucsd.edu} 
\icmladdress{Computer Science and Engineering, UCSD} % PUT KEYWORDS HERE FOR SUBMISSION
%
% The author names and address list should only appear in the accepted version. 
% \icmlauthor{Your Name}{email@yourdomain.edu}
% \icmladdress{Your Fantastic Institute,
%             314159 Pi St., Palo Alto, CA 94306 USA}
% \icmlauthor{Your CoAuthor's Name}{email@coauthordomain.edu}
% \icmladdress{Their Fantastic Institute,
%             27182 Exp St., Toronto, ON M6H 2T1 CANADA}

\vskip 0.3in
]

\begin{abstract} 
We present a new boosting algorithm, motivated by the large margins
theory for boosting. We give experimental evidence that the new
algorithm is significantly more robust against label noise than 
existing boosting algorithm. 
\end{abstract} 

\section{Introduction}
Since the invention of Adaboost by Freund and
Schapire~\yrcite{FreundSc96,FreundSc97} it has become very popular
with both theoreticians and practitioners of machine learning. Many
variants of the algorithm have been devised.

One of the most intriguing properties of Adaboost is the fact that it
tends not to overfit. In many cases the test error of the generated
classifier continues to decrease even after the training error has
decreased to zero~\cite{DruckerCo96,Quinlan96,Breiman98,SchapireFrBaLe98}. There
are two main theories for explaining this behaviour. The first is the
large margins theory, proposed by Schapire et
al~\yrcite{SchapireFrBaLe98}. This theory is closely related to the
theory of support vector machines (SVM)~\cite{CortesVa95}. The focus
of large margin theory is on the task of minimizing the {\em
  classification} error rate on the test set. The different theory,
proposed by Friedman et al~\yrcite{FriedmanHaTi00}, related Adaboost to
logistic regression. The main focus of this theory is on maximizing
the likelihood of a conditional probability distribution represented
as a logistic function. The decrease in classification error is seen
as a by-product of the increase in the likelihood.

One problem with Adaboost that has been realized early on is it's
sensitivity to noise~\cite{Dietterich00}. The performance of Adaboost
deteriorates rapidly when random label noise is added to the training
set.  Friedman et al proposed a variant of Adaboost, which they named
gentle Adaboost or Logitboost, which is significantly better than
Adaboost at tolerating label noise. Similar algorithms to Logitboost
are log-loss Boost proposed by Collins, Schapire and
Singer~\yrcite{CollinsScSi02} and MAdaboost, proposed by Domingo and
Watanabe~\cite{DomingoWa00a}, as these algorithms are very similar to
Logitboost, we will refer only to Logitboost from now on. While
Adaboost puts unbounded large weights on mislabeled examples, the
weight placed on any example by Logitboost is bounded. This decreases
the penalty on mislabeled examples and increases the ability of the
algorithm to tolerate noise.

All of these algorithms can be described as methods for minimizing a
potential function using gradient
descent~\cite{MasonBaBaFr99b}. Moreover, the potential function used
by Adaboost, Logitboost, Logloss Boost and MAdaboost are all
convex. The minimum of convex potential functions can be computed
efficiently, which is the reason boosting is an efficient algorithm.
However, Long and Servedio~\yrcite{LongSe08} prove that any boosting
algorithm that is based on a convex potential function can be defeated
by random label noise. They present a simple construction of a
distribution that cannot be learned using such algorithms.

In this paper we present a new boosting algorithm, which we call
Robustboost, which is significantly more robust against label noise
than either Adaboost or Logitboost. The new algorithm is based on the
Freund's Boost-by-Majority algorithm~\yrcite{Freund95} and
Brownboost~\yrcite{Freund01}. The algorithm is a potential based
algorithm. However, the potential function is {\em not} convex and it
changes during the boosting process.

The paper is organized as follows. In Section~\ref{sec:potential} we
describe the potential based approach for learning linear
discriminators and the problems associated with label noise and convex
potential functions. In Section ~\ref{sec:margins} we discuss the
margin based explanation for Adaboost's resistance to overfitting and
how to apply it to problems in which the data is not linearly
separable.  In Section~\ref{sec:algorithm} we present the new boosting
algorithm. In Section~\ref{sec:experiments} we give the experimental
evidence that the new algorithm has superior robustness against label
noise. We conclude in Section~\ref{sec:conclusion}.

\section{Learning Linear Discriminators under noise}
\label{sec:potential}
To simplify this explanation, we fix the set of base classifiers and
assume that the weak learner picks the base classifier with the
smallest weighted error at each iteration. In this section we focus on
the the problem of minimizing the number of mistakes that the combined
classifier makes on the training set. The performance of the learned
classifier on examples outside the training set will be discussed in
Section~\ref{sec:margins}.

\newcommand{\h}{\vec{h}}
\newcommand{\avec}{\vec{\alpha}}

We assume that we are given a training set $S=\langle
(x_1,y_1),\ldots,(x_N,y_N) \rangle$ where $x_i$ are the feature
vectors and $y_i \in \{-1,+1\}$ are the binary labels.  The
classification rule generated by Adaboost is of the form $c(x) =
\sign(\sum_{i=1}^d \alpha_i h_i(x))$, where $\alpha_1,\ldots,\alpha_d$
are real numbers and $h_1,\ldots,h_d$ is the fixed set of base
rules. We assume that the range of the base classifiers is
$\{-1,+1\}$. As the base classifiers are fixed, we can represent the
feature vector $x$ by the $d$ dimensional binary vector $\h=\langle
h_1(x),\ldots,h_d(x) \rangle$. This reduces the problem to that of
learning a weights vector $\avec=\langle \alpha_1,\ldots,\alpha_d
\rangle$ that defines a good linear discriminator of the form
$\sign(\avec \cdot \h)$.

We call the argument of the signum function the {\em score function}
$s(x) = \avec \cdot \h$. The (un-normalized) {\em margin} of an
example $(x,y)$ is defined to be the product of the score function and
the label $m(x,y)=ys(x)$. Clearly $m(x,y)>0$ if and only if the
classification $c(x)$ is correct. Thus the indicator function
$\IndF{m(x_i,y_i) \leq 0}$ is $1$ if $c(x)=y$ and $0$ otherwise, we call
this indicator function the ``error step function''.

Our goal in this section is to find the linear discriminator which
minimizes the training error, which can be expressed as
\begin{equation} \label{eqn:trainingError}
P_S[c(x) \neq y] = \frac{1}{N}\sum_{i=1}^N \IndF{m(x_i,y_i) \leq 0}
\end{equation}
If there exists a linear classifier whose training error is zero we
say that the training data is {\em linearly separable}. In this case
there are several provably efficient algorithms for finding the
separating hyperplane (for example, the perceptron algorithm). On the
other hand, when the training data is not linearly separable there is
no known efficient algorithm for finding the vector $\avec$ that
minimizes the training error. In fact, it is known that this is an
NP-hard, and that it is also NP-hard to find the linear classifier
with the minimal number of disagreements or to approximate it within a
factor of $2-\epsilon$~\cite{FeldmanGoKhKu06}.

\begin{figure}[t]
%\vskip -0.2in
\begin{center}
\centerline{\includegraphics[width=\columnwidth]{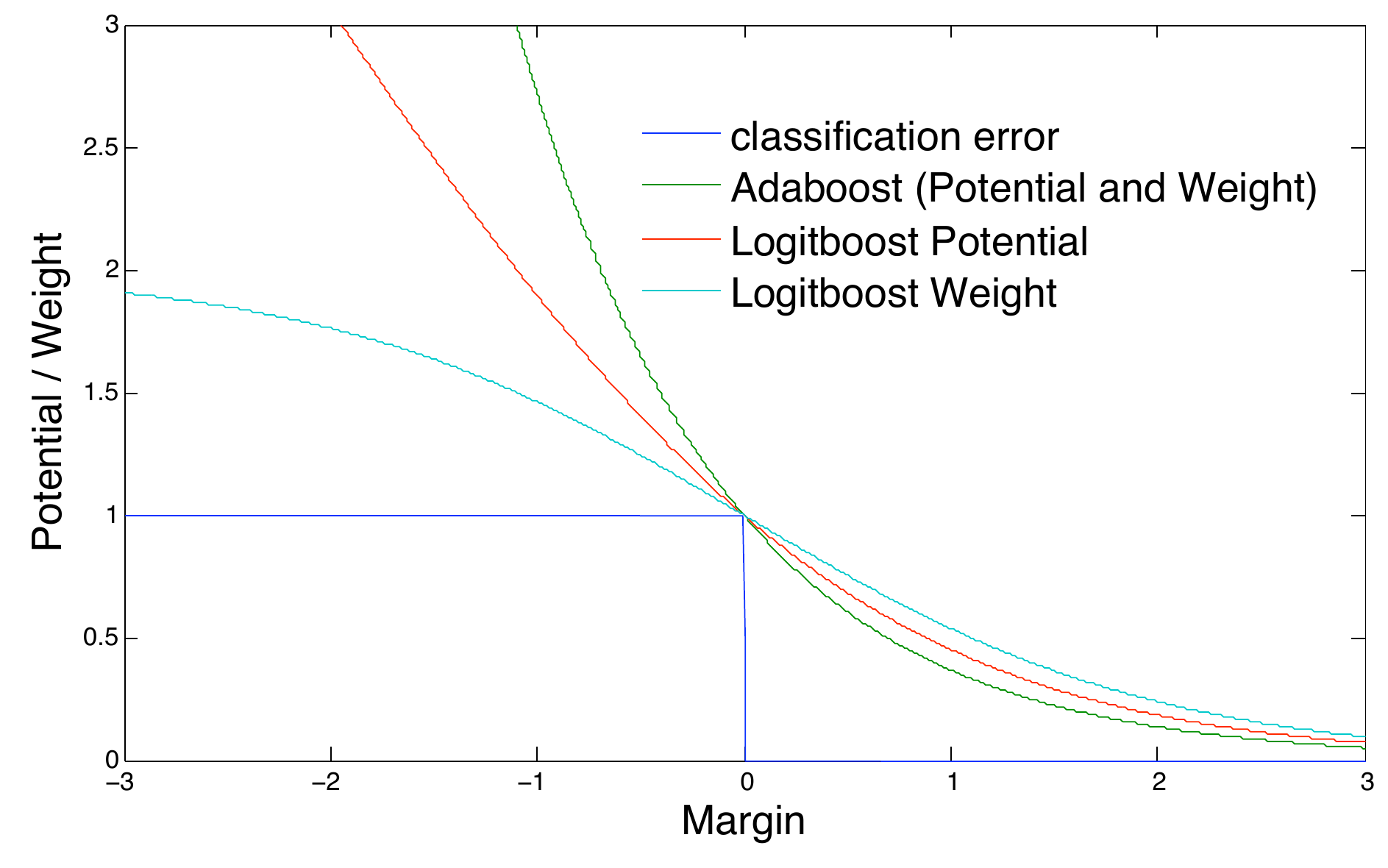}}
\caption{Potential and weight functions for Adaboost and Logitboost}
\label{fig:lossFunctions}
\end{center}
\vskip -0.4in
\end{figure} 

As finding good linear separators is a problem of great practical
importance algorithms have been developed that can find the optimal
separating hyperplane under particular assumptions regarding the
distribution of the examples. Two prominent examples are linear
discriminant analysis, which assumes that the two classes are normally
distributed with equal covariance matrices and logistic regression
which assumes that the conditional probability of the label given the
input is described by the logistic function.  Friedman et
al~\yrcite{FriedmanHaTi00} show that Logistic regression is closely
related to Adaboost and to Logitboost.

Logitboost and Adaboost can be represented using a potential
function. The potential function $\Phi$ is a decreasing function of
the margin $m$ which upper bounds the error step function $\Phi(m)
\geq \IndF{m \leq 0}$. As the average potential is an upper bound on
the classification error, decreasing it is a good heuristic for
decreasing the classification error. The potential function for
Adaboost is $\exp(-m)$ and the potential function for Logitboost is
$\ln(1+\exp(-m))$. These functions are very close when $m>0$ but
diverge when $m$ is negative because for $m \ll 0$ the potential for
Logitboost is approximately linear $\ln(1+\exp(-m)) \approx
-m$. Minimizing these potential functions can be done very effectively
using gradient descent methods. In particular using the chain rule we
get a simple expression for the derivative of the average potential
w.r.t. $\alpha_i$. Denoting $m(x_j,y_j)$ by $m_j$ we get
\begin{eqnarray*}
\lefteqn{\frac{d}{d \alpha_i} \frac{1}{N} \sum_{j=1}^N \phi(m_j) = 
\frac{1}{N} \sum_{j=1}^N \frac{d m_j}{d \alpha_i}
\left. \frac{d\phi(m)}{dm} \right|_{m=m_j}} &&\\
&=& 
\frac{1}{N} \sum_{j=1}^N y_j h_i(x_j) 
\left. \frac{d\phi(m)}{dm} \right|_{m=m_j}
\end{eqnarray*}
It is therefor natural to define a {\em weight function} $w(m)$ that is
(minus) the derivative of the potential function with respect to
$m$. Using this notation we get the expression
\[
\frac{d}{d \alpha_i} \frac{1}{N} \sum_{j=1}^N \phi(m(x_j,y_j)) = 
- \frac{1}{N} \sum_{j=1}^N y_j h_i(x_j) w(m(x,y))
\]
which has the attractive interpretation that the derivative of the
average potential w.r.t. $\alpha_i$ is equal to correlation between
$h_i(x)$ and $y$ over the training examples, weighted by
$w(m(x,y))$. The weights represent the relative importance of
different examples in reducing the average potential. The weight
function for Adaboost is $w(m)=\exp(-m)$ and the weight function for
Logitboost is $w(m) = 1/(1+\exp(m))$. Note that the weights assigned
by Adaboost rapidly increase to infinity when $m<0$ while the weights
assigned by Logitboost are at most $1$.

Consider now what happens when we apply Adaboost or Logitboost to a
linearly separable dataset to which we added independent label
noise. In other words, we take a training set $S$ which is perfectly
classified with the a hyper-plane defined by $\avec$ and flip the
label of each example independently at random with probability
$q<1/2$. The training set is no longer linearly separable. However,
the classifier $\alpha$ which previously separated the two classes is
still almost optimal as its error rate is approximately $q$ which is
the lowest achievable error rate. On the other hand $\alpha$ is
unlikely to be the point at which the average {\em potential} of
Adaboost or Logitboost achieves it's minimum. Long and
Servedio~\yrcite{LongSe08} give a rigorous proof of this fact. Here we
give a short intuitive explanation.

If there are examples in the clean dataset whose margin $m$ with
respect to $\avec$ is large, then a fraction of about $q$ of these
examples now have a margin $m$ that is a large {\em negative}
number. If the margin of such a noisy example is $m \ll 0$ then the
potential assigned to it by Adaboost is $\exp(-m)$ and the potential
assigned to it by Logitboost is about $-m$. Both potentials are larger
than the error step function which they bound (see
Figure~\ref{fig:lossFunctions}). Moving away from $\avec$ can decrease
the distance between the misclassified examples and the decision
hyperplane, i.e. increase the corresponding negative margins, thereby
decreasing the potential of those examples and therefor the overall
average potential. The potentials and the weights assigned to
mislabeled examples by Adaboost are much larger than those assigned by
Logitboost. Therefor Logitboost is much more robust against label
noise than Adaboost. However, Long and Servedio show that random label
noise is a problem for any boosting algorithm that uses a {\em convex}
potential function. The algorithm we propose in this paper is based on
a non-convex potential function. It thus even more robust the
Logitboost against label noise. In Section~\ref{sec:LongExperiments}
we give an experimental evidence that our algorithm is much more
robust than Logitboost and Adaboost for the classification problem
suggested by Long and Servedio.

\section{The large margins theory}
\label{sec:margins}

In the previous section we focused on minimizing the training
error. However, it is clear that Adaboost and Logitboost are doing
something more than minimizing training error. In many
experiments~\cite{DruckerCo96,Quinlan96,Breiman98,SchapireFrBaLe98}
the test error of the generated strong classifier continues to
decrease for many boosting iterations {\em after} the training error
becomes zero. Characterizing the criterion that Adaboost optimizes
which is a better predictor of the test error than the training error
is an important step towards finding a better boosting algorithm. To
this end we use the large margins theory of Schapire et
al~\yrcite{SchapireFrBaLe98}.

\newcommand{\nm}{\bar{m}}
We use the following terminology. Recall that the definition of the
un-normalized margin is $m(x,y) = y \sign(\sum_i \alpha_i h_i(x))$.
We define the {\em normalized} margin to be 
$ \nm(x,y) = \paren{y \sign(\sum_i \alpha_i h_i(x))}/\paren{\sum_i |\alpha_i|} $

The {\em generalization error} of a the classifier $c(x)$ is the
probability that $c(x) \neq y$ when $(x,y)$ is generated by the
underlying distribution $D$. Using the margin notation we express the
generalization error as $P_D[\nm(x) \leq 0]$. We denote the optimal
classification error by $c^*$.

The {\em margin theory} posits that large positive margins on
training examples are predictive of small generalization
error. Specifically, Theorem 1 in~\cite{SchapireFrBaLe98} states that
for any $\theta>0$, with probability $1-\delta$ over the random choice
of the training set
\begin{eqnarray}
\lefteqn{P_D[\nm(x) \leq 0] \leq P_S[\nm(x) \leq \theta]} && \label{eqn:margins-bound}\\
&+&O\left( \frac{1}{\theta} \sqrt{\frac{\log N \log d +
      \log(1/\delta)}{N}} \right). \nonumber
\end{eqnarray}
where $P_D[m(x) \leq 0]$ is generalization error, i.e.  the
probability of making a mistake with respect to the underlying
distribution, $P_S[\nm(x) \leq \theta]$ is the fraction of the examples
in the training set for which the margin is smaller than $\theta$, $d$
is the number of base rules and $N$ is the size of the training set.

The interpretation of this theorem is that to minimize the
generalization error we should find linear classifier that minimizes
the number of training examples for which $\nm(x) \leq \theta$ for a
large value of $\theta$. Note that varying $\theta$ has opposite
effects on the two term in the bound. Increasing $\theta$ causes
the first term to increase to $1$ while decreasing $\theta$ towards
$0$ causes the second term to blow up. 

Schapire et al show that Adaboost will tend to decrease the bound
given in Equation~(\ref{eqn:margins-bound}) by increasing the value of
$\theta$ for which the term $P_S[\nm(x) \leq \theta]$ is equal to
zero. In other words, by maximizing the {\em minimal} margin.
Breiman~\cite{Breiman97b} and Grove and Schuurmans~\yrcite{GroveSc98}
give experimental evidence against this explanation for why Adaboost
does not overfit. They maximized the minimal margin directly and
showed that this does not tend to decrease the generalization error.

However, setting the first term in Equation~\ref{eqn:margins-bound} to
zero and minimizing the second term is often not the best way to minimize
the bound. If the training set is not linearly separable it is
impossible to set the term $P_S[\nm(x) \leq \theta]$ to zero for any
$\theta \geq 0$. We can still get meaningful bounds from
Equation~\ref{eqn:margins-bound}, but in order to do that we need an
algorithm for finding a weight vector $\avec$ for which $\nm(x_i,y_i)
> \theta$ for {\em most} but not all of the training examples.

We therefor redefine the goal of the boosting algorithm, instead of
minimizing the number of mistakes on the training set, we define the
goal as minimizing the number of examples whose normalized margins is
smaller than some value $\theta>0$. Stated using our defined notation,
our goal is to minimize 
\begin{equation} \label{eqn:margin-goal-function}
(1/N) \sum_{i=1}^N \IndF{\nm(x_i,y_i) \leq \theta}.
\end{equation}
In the next section we describe the potential-based boosting algorithm
for minimizing this target function.

\section{Robustboost}
\label{sec:algorithm}

Our proposed algorithm, which we call Robustboost, is a variation on
Brownboost algorithm proposed by Freund~\yrcite{Freund95} which, in
turn, is based on Freund's Boost-by-Majority (BBM)
algorithm~\yrcite{Freund01}. We give a brief description of
Boost-by-majority and Brownboost and then describe Robustboost.

BBM is based on the idea of {\em finite horizon}. The number of
boosting iterations is set in advance based on an {\em error goal}
parameter $\epsilon$ which is given to BBM as input. While all
boosting algorithm give small weight to examples with large positive
margins, BBM gives small weight to examples with large {\em negative}
margins on the later iterations. Intuitively, it ``gives up'' on
examples which are so far on the incorrect side of the boundary that
they are unlikely to be classifier correctly at the end. These
examples become part of the training error, which is quantified by
$\epsilon$. As the weight for a given margin is the derivative of the
potential, this implies that the potential has decreasing slope for
large positive margins as well as large negative margins. In other
words, the potential function is non-convex.

One deficiency of BBM is that it is not adaptive. In other words, the
base classifier added at each iteration is assigned a weight of one
regardless of its accuracy (a base classifier {\em is} assigned a
larger weight if it is added in multiple iterations).
Brownboost is the adaptive version of BBM. It assigns based
classifiers with small error larger weight than base classifiers with
error close to 1/2. Brownboost uses a real valued variable called
``time'' and denoted by $t$. Before the first iteration $t=0$ and it
is increased in each iteration. While BBM terminates after a
pre-defined number of iterations, Brownboost terminates when $t$
reaches a pre-defined value $c$. The parameter $c$ defines the horizon
of the boosting process and is pre-computed according to the target
error $\epsilon$. As $\epsilon \to 0$, $c \to \infty$ and Brownboost
becomes equivalent to Adaboost. In other words, Adaboost is a special
case of Brownboost where the target error is zero.

The algorithm we propose here is very similar to Brownboost. The main
difference is that instead of minimizing the training error it's goal
is to minimize the margin-based cost function defined in
Equation~(\ref{eqn:margin-goal-function}). In order to minimize
this cost function it needs to use a normalized weight
vector. Designing an algorithm that would keep the $L_1$ norm of the
weight vector bounded proved difficult. Our solution is to normalize
the weight vector so that the variance of the scores is bounded. In
other words, the algorithm controls the weight vector $\avec$ so that 
$\mbox{Var}(\avec \cdot h)$ is small. This is achieved by adding to
the drift in the underlying Brownian motion process a component which
pushes the examples towards zero. This makes the underlying process
equivalent to the mean-reverting Ornstein-Uhlenbeck process (see page
75 in~\cite{Oksendal2003}). 

\begin{figure}[t]
%\vskip -0.2in
\begin{center}
\centerline{\includegraphics[width=\columnwidth]{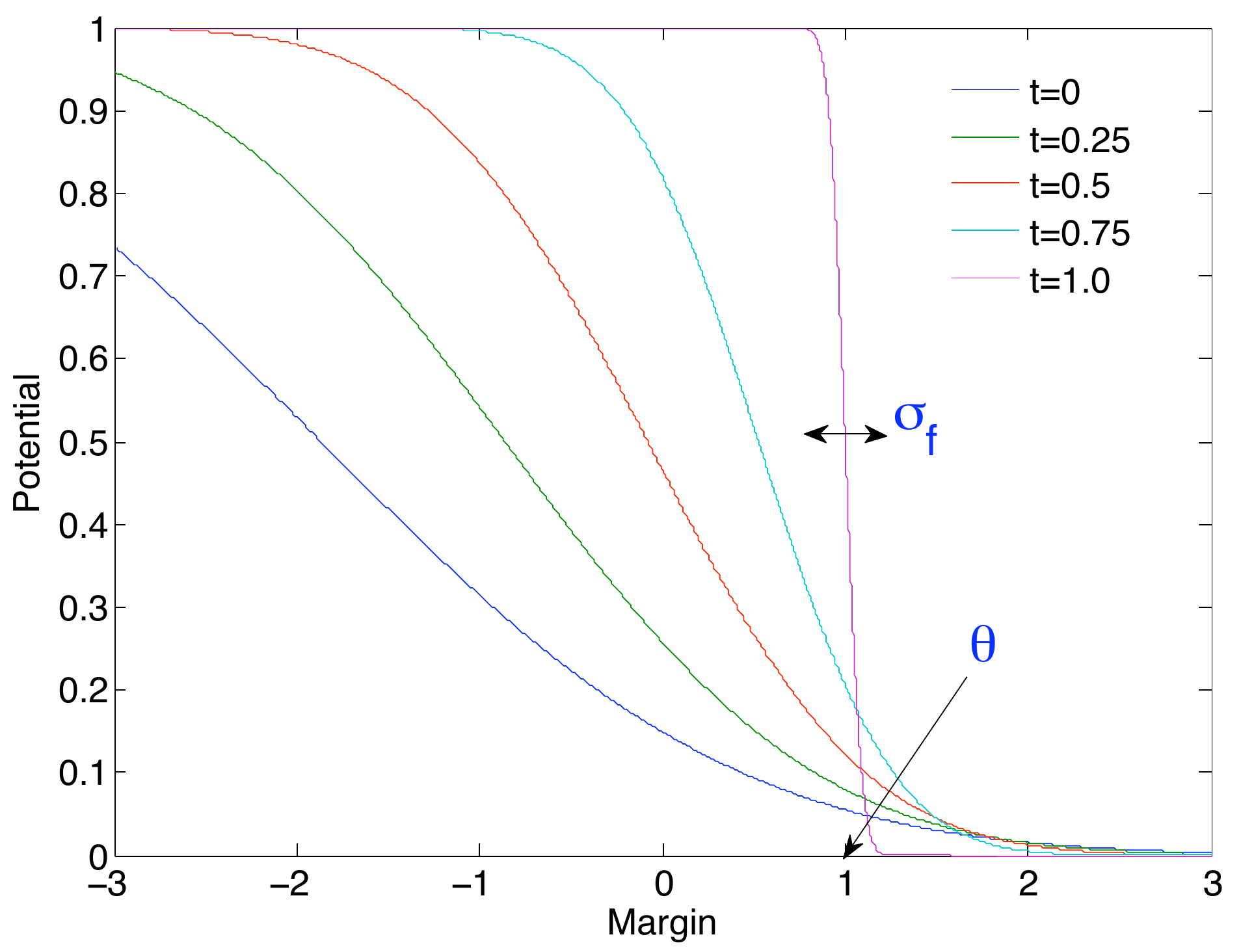}}
\caption{The potential functions used by Robustboost as a function of
  $t$. The potential function at $t=1$ is a close approximation of the
  margins error function $\IndF{\nm(x_i,y_i) \leq \theta}$.}
\label{fig:RobustBoostPotentials}
\end{center}
\vskip -0.4in
\end{figure} 

We now describe the details of Robustboost. In our setup the range of
the time variable is $0 \leq t \leq 1$. Denoting the margin by $\s$ we
define the potential function to be 
\begin{equation} \label{eqn:NormBoostConfRated-pot}
\rpot{\s}{t}=
1-\erf \paren{ \frac{\s - \mu(t)}{\sigma(t)}}.
\end{equation}
Where $\erf$ is the error function
\[
\erf(a) \doteq \frac{1}{\sqrt{\pi}} \int_{-\infty}^a e^{-x^2} dx
\]
$\mu(t)$ and $\sigma(t)$ are defined by the equations
\begin{equation} \label{eqn:sigma-evolution}
\sigma^2(t) = (\sigmaf^2+1) e^{2(1-t)} -1
\end{equation}
and
\begin{equation} \label{eqn:mu-evolution}
\mu(t) = (\theta-2\rho) e^{1-t} +2\rho
\end{equation}
Where $\theta,\sigmaf$ and $\rho$ are parameters of the algorithm that
we will describe shortly. Taking the partial derivative of
$\rpot{\s}{t}$ with respect to $\s$ we get that the weight function is
\begin{equation} \label{eqn:NormBoostConfRated-wt}
\rw{\s}{t} =
  \exp \paren{-\frac{\paren{\s-\mu(t)}^2}{2\sigma(t)^2}} 
\end{equation}

The parameter $\theta \geq 0$ is the goal margin, as defined in
Equation~(\ref{eqn:margin-goal-function}). It is set using
cross-validation. Increasing $\theta$ decreases the difference between
the performance on the training set and on the validation set. The
parameter $\sigma_f>0$ defines the slope of the step in the final
potential function (see Figure~\ref{fig:RobustBoostPotentials}). We
set $\sigmaf=0.1$ to avoid numerical instability when $t$ is close to
1.

Like Brownboost, Robustboost is a self-terminating algorithm. It
terminates when $t \geq 1$. If the error goal $\epsilon$ is set too
small then Robustboost will not terminate. Setting the right value for
$\epsilon$ is done by searching for the minimal value of $\epsilon$
for which the algorithm terminates within a reasonable number of
iterations.  Setting $\epsilon$ determines the value of the parameter
$\rho$. That value is the solution to the following equation
\begin{equation} \label{eqn:initial-potential-confRated}
\epsilon = \rpot{0}{0} = 
1-\erf 
  \paren{2(e-1)\rho - e \theta \over \sqrt{e^2(\sigmaf^2+1)-1}}
\end{equation}

~\\

{\bf Robustboost}
%%%%%%%%%%%%%%%%%%%%%% BEGIN Algorithm Description
%\begin{algorithm}
%   \caption{Robustboost}
%   \label{alg:Robustboost}
\begin{algorithmic}
\STATE {\bfseries parameters:} $\epsilon>0$, $\theta>0$, $\sigmaf>0$
\STATE {\bfseries training set:} $(x_1,y_1),\ldots,(x_\N,y_\N)$  where
$x_j \in \X$, $y_j \in \{-1,+1\}$
\STATE
\STATE {\bfseries set} $\rho$ to satisfy Equation~(\ref{eqn:initial-potential-confRated}).
\STATE {\bfseries initialize} $k=1$, $t_1:=0$, $\hf_0 \equiv 0$
\STATE $\s(1) := 0, \ldots \s(\N):=0$
\STATE
\REPEAT
\STATE {\bfseries Define} the distribution $D_k$ over the $\N$ training examples by normalizing
$\rw{\s}{t}$ defined in Equations~(\ref{eqn:NormBoostConfRated-wt},\ref{eqn:sigma-evolution},\ref{eqn:mu-evolution})
\[
D_k(j) = \frac{\rw{\s(j)}{t_k}}{Z},\;\; Z = \sum_{j=1}^\N \rw{\s(j)}{t_k}
\]
\STATE {\bfseries call} base learner and get $h_k:\X\rightarrow \{-1,+1\}$ which is slightly
 correlated with the label:
\[ \Exp{j \sim D_{k}}{y_j h_k(x_j)} > 0 \]
\STATE {\bfseries find} $\dels_k>0,1-t_k \geq \deltat_k>0$ that simultaneously 
satisfy the following two equations:
\[
\sum_{j=1}^{\N} y_j h_k(x_j) \rw{\s'(j)}{t_k+\deltat_k} = 0
\mbox{~or~}
\deltat_k = 1-t_k
\]
\[
\sum_{j=1}^{\N} \rpot{\s(j)}{t_k} = \sum_{j=1}^{\N} \rpot{\s'(j)}{t_k+\deltat_k}
\]
where 
\[
\s'(j) \doteq \s(j)e^{- \deltat_k}+y_j h_k(x_j)\dels_k
\]
and $\rpot{\cdot}{\cdot},\rw{\cdot}{\cdot}$ are defined by
Equations~(\ref{eqn:NormBoostConfRated-pot},\ref{eqn:NormBoostConfRated-wt},\ref{eqn:sigma-evolution},\ref{eqn:mu-evolution}).
\STATE {\bfseries update:} $t_{k+1}:=t_k+\deltat_k$
$\forall 1 \leq j \leq  \N\;\;\; \s(j) := \s'(j) $
$\hf_k = \hf_{k-1} e^{-\deltat_k} + \dels_k h_k$
\STATE $k:=k+1$
\UNTIL $t_{k} \geq 1$.
\STATE {\bfseries Output:} the final hypothesis $\hf_k$.
\end{algorithmic}
%\end{algorithm}
%%%%%%%%%%%%%%%%%%%%%% END Algorithm description

\iffalse
\begin{algorithm}[tb]
   \caption{Bubble Sort}
   \label{alg:example}
\begin{algorithmic}
   \STATE {\bfseries Input:} data $x_i$, size $m$
   \REPEAT
   \STATE Initialize $noChange = true$.
   \FOR{$i=1$ {\bfseries to} $m-1$}
   \IF{$x_i > x_{i+1}$} 
   \STATE Swap $x_i$ and $x_{i+1}$
   \STATE $noChange = false$
   \ENDIF
   \ENDFOR
   \UNTIL{$noChange$ is $true$}
\end{algorithmic}
\end{algorithm}
\fi

\section{Experiments}
\label{sec:experiments}

We report the results of two sets of experiments using synthetic
data distributions. The first distribution is taken from Long and
Servedio~\cite{LongSe08}. The second is taken from Mease and Wyner~\yrcite{WynerMe07}.

\subsection{The Long/Servedio problem}
\label{sec:LongExperiments}

Long and Servedio suggested the following challenging classification
problem. The input is a binary feature vector of length 21: $\langle
x_1,\ldots,x_{21} \rangle$ where $x_i \in \{-1,+1\}$ and the output is
$y \in \{-1,+1\}$. A random example is generated as follows. First,
the label $y$ is chosen with equal odds to be $-1$ or $+1$. Given $y$
the features $x_i$ are generated according to the following mixture
distribution:
\begin{itemize}
\item {\bf Large margin examples}: With probability $1/4$ all of the
  $x_i$ are set equal to the label $y$.
\item {\bf Pullers}: With probability $1/4$ we set the first eleven
  coordinates equal to the label $x_1=x_2=\cdots=x_{11}=y$ and
  $x_{12}=x_{13}=\cdots=x_{21} = -y$.
\item {\bf Penalizers}: With probability $1/2$ we do the
  following. Choose at random 5 coordinates out of the first 11 and 6
  coordinates out of the last 10 and set those equal to the label $y$.
Set the remaining 10 coordinates to $-y$.
\end{itemize}
 
It is easy to check that the majority vote rule $\sign(\sum_i x_i)$ is
a perfect classifier for this data. To learn this classifier using
boosting we define the base classifiers to be single coordinates,
i.e. $h_i(\vec{X}) = x_i$. Clearly, using these base classifiers the
target classifier can be represented exactly. As this is a linearly separable
distribution, both Adaboost and Logitboost can learn it perfectly.

However, if we add label noise to this problem, i.e. if we flip each
label with probability $0.1$ then the performance of both Adaboost and
Logitboost degrades severely. The optimal rule remains the majority
vote rule, but the average potential of this rule is very large and
the minimal potential is achieved by a sub-optimal classifier.  Long
and Servedio show that any potential based boosting algorithm that
uses a convex potential function will fail to find a classifier whose
training error is close to that of the majority vote rule. It is
important to note that the failure of these learning algorithms
demonstrates itself already on the training set and is a problem of
{\em under-}fitting, not overfitting, the training data.

We compare the performance of Robustboost, Logitboost and Adaboost on
this problem. We run each boosting algorithm for 300 iterations. We
generated 10 datasets, each consisting of 800 training examples. We
run each algorithm on each of the data sets and compute the training
error for each case. We also compute the error with respect to the
{\em clean} labels, which have not been corrupted by noise. We report
the average and standard deviation of these errors and their relative
order for individual datasets.

We set the parameters of Robustboost to be $\sigmaf=0.1$ and
$\epsilon=0.14$. The value of $\epsilon$ was chosen so that that
Robustboost terminates after 200-300 iterations for most datasets. We
tried two settings $\theta$. In one setting $\theta=0$, i.e. the goal
of the algorithm is to minimize the training error. In the other
setting $\theta=0.2$ which means that the goal of the algorithm is to
minimize the number of training examples whose margin is less than
$0.2$. The numbers in this and the following tables corresponds to
percent errors (i.e. there are numbers between 0 and 100).

\begin{small}
\begin{tabular}{|l|c|c|c|c|}
\hline
  & Ada & Logit & Robust    & Robust \\
  &     &        & $\theta=0$& $\theta=0.2$ \\
\hline
Err & $28.1 \pm 1.5$ & $26.6 \pm 1.5$ & $13.2 \pm 1.3$ & $10.0 \pm 1.3$ \\
\hline
Clean & $24.3 \pm 1.7$ & $22.6 \pm 1.7$ & $5.5 \pm 2.5$ & $0.9 \pm 1.0$ \\
\hline
\end{tabular}
\end{small}

The relative order of the errors was always the same. The error of
Robustboost is far smaller than that of Logitboost, which is slightly
better than that of Adaboost. Surprisingly, the error of Robustboost
is further improved when we set $\theta=0.2$. The difference is even
more pronounced when comparing the predictions of the classifier to
the noiseless labels. In particular, the error of Robustboost is about
$1.0\%$ even though the data on which it was trained has $10\%$ error
with respect to the clean data. In other words, Robustboost is able to
detect and correct most of the mislabeled examples.

\begin{figure}[t]
%\vskip 0.2in
\begin{center}
\centerline{\includegraphics[width=\columnwidth]{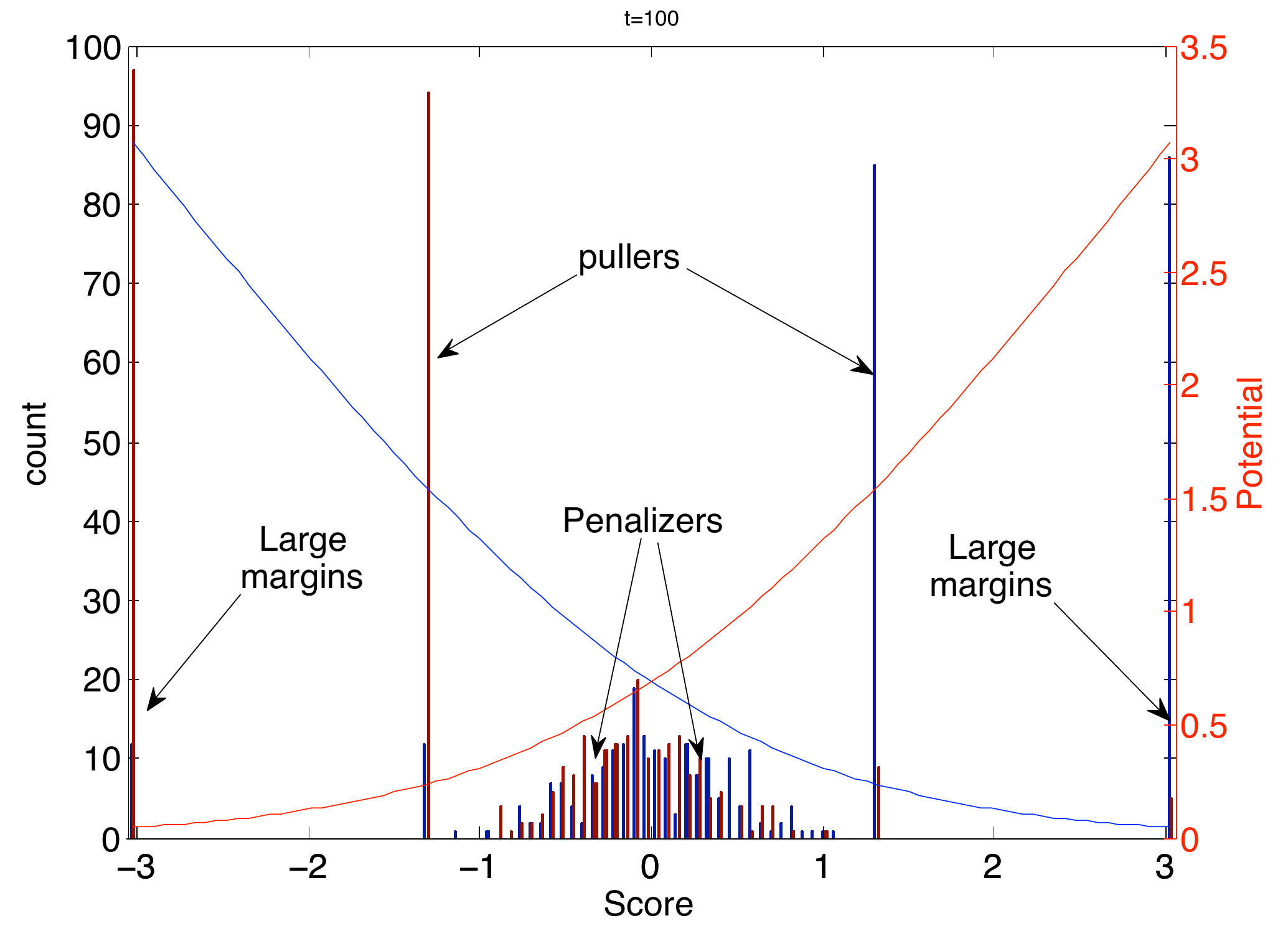}}
\caption{The score distribution and the potential functions for
  Logitboost on a Long/Servedio training set on iteration 100.}
\label{fig:LongGentleBoost}
\end{center}
\vskip -0.4in
\end{figure} 

We can gain insight into the reasons that Robustboost succeeds while
Adaboost and Logitboost fail by comparing the evolution of the score
distributions for the different algorithms. In
figure~\ref{fig:LongGentleBoost} we show the score distribution to
which Logitboost converges after 100 iterations, this distribution
changes very little from iteration 100 to iteration 300. What we see
is that the algorithm converged to a minimal potential vector $\avec$
in which the large margin examples and the pullers are well separated,
but the penalizers are distributed more or less randomly. The reason
is that the mislabeled large margins and pullers have relatively large
weights (the derivative of he potential is close to one) while the
weight of each individual penalizer is small. As the penalizers are
sparse, they cannot ``pull'' $\avec$ from the direction suggested by
the pullers and large margin examples and so about half 0f them are
mislabeled, contributing about 25\$ to the training error.

\begin{figure}[t]
\begin{center}
\centerline{\includegraphics[width=\columnwidth]{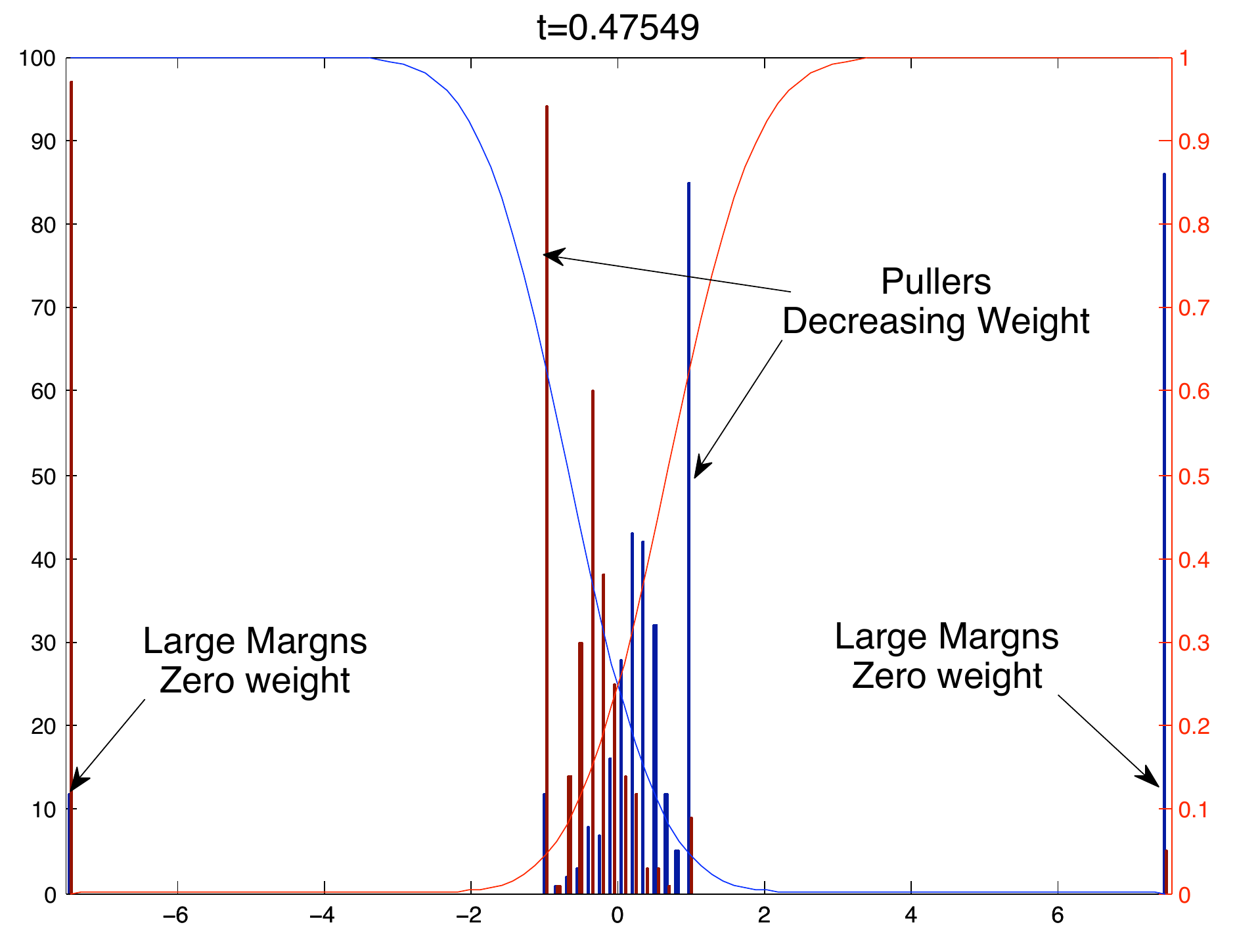}}
\centerline{\includegraphics[width=\columnwidth]{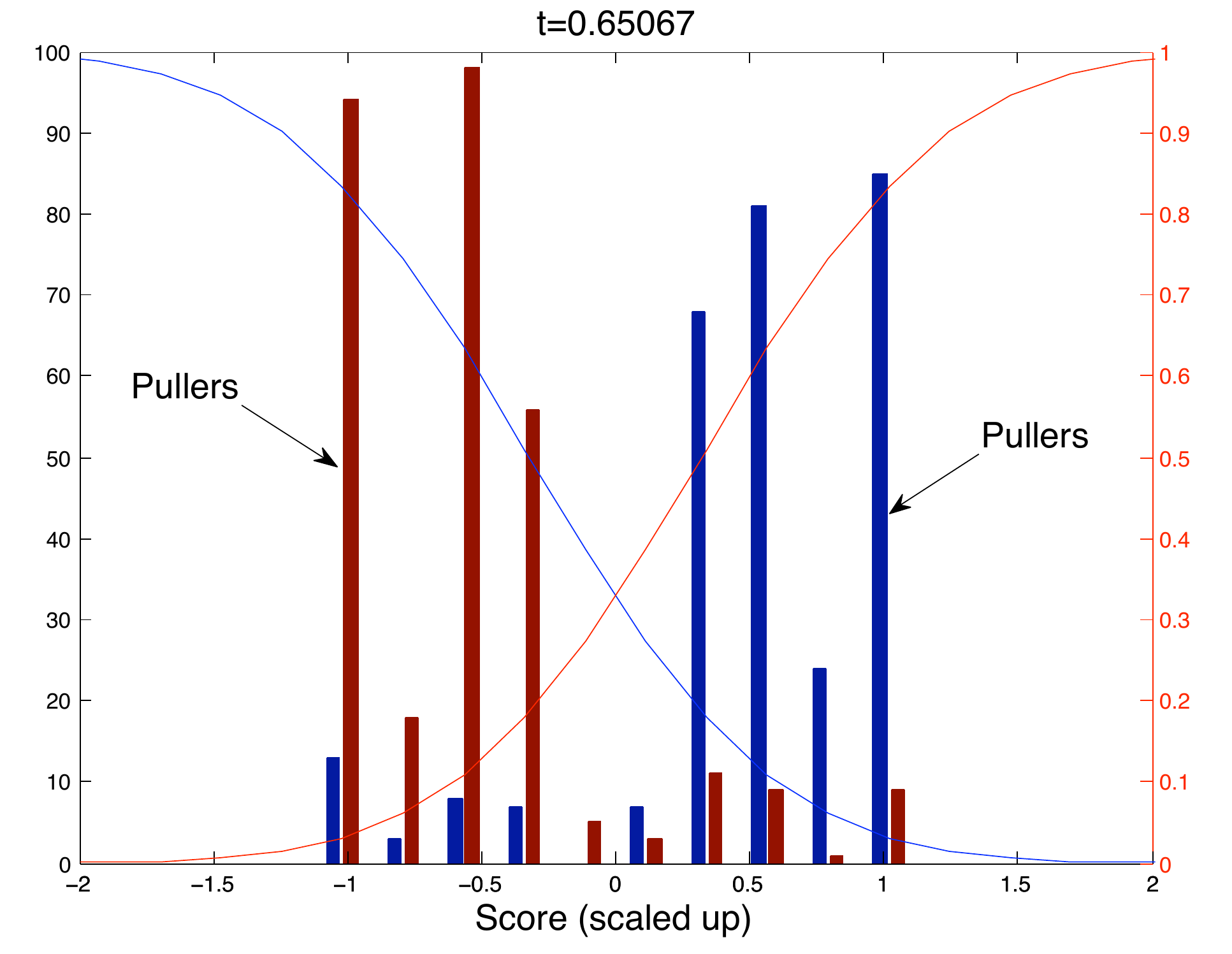}}
\caption{The score distribution and the potential functions for
Robustboost on a Long/Servedio training set on iterations 100 (top)
and 200 (bottom).}
\label{fig:LongRobustBoost}
\end{center}
\vskip -0.4in
\end{figure}

Contrast this with the score distributions shown in
Figure~\ref{fig:LongRobustBoost}. After 100 boosting iterations the
potential is such that the weight of large margin examples is close to
zero {\em whether or not they are mislabeled} and the weight of
mislabeled pullers is smaller than it was with Logitboost. This means
that the algorithm ignores the large margin examples and concentrates
on the pullers and the penalizers, without giving the pullers too much
weight. The result is that after 200 iterations many of the penalizers
are classified correctly and the pullers are mixed in with the
penalizers. Note that for the ideal solution the margins of pullers
and penalizers that are not mislabeled is equal to $11/21 - 1/2 = 1/42$.
Our main point is that Robustboost avoids the incorrect mini ma that
trap Adaboost and Logitboost by ignoring examples with large negative margins.

\subsection{The Mease/Wyner problem}
In this section we report results of experimental comparisons using
synthetic distributions analyzed by Mease and Wyner~\yrcite{WynerMe07}. In this
case a majority vote over the base classifier can only approximate the
target classifier which significantly complicates the problem.

The input to this classification problem is a $d=20$ dimensional
vector $\vec{x}$ where each coordinate of $x$ is chosen IID according
to the uniform distribution over the segment $[0,1]$. The label $x$ is
$1$ if $\sum_{i=1}^5 x_i \geq 2.5$ and $-1$ otherwise. The base
classifiers we tested are decision stumps, i.e. rules of the form
$\sign(x_i - a_i)$ or a 2 level decision tree made out of these
decision rules. Unlike the Long and Servedio distribution, a finite
number of these base classifiers cannot exactly represent the target
rule. Mease and Wyner use this distribution to compare the effects of
random label noise on Adaboost and Logitboost. We add Robustboost to
the comparison.

In each experiment we use $2000$ training examples and $2000$ test
examples (we tried $200$ examples but the between-experiment variation
was too large to draw significant conclusions). We repeat each
experiment 15 times and report the mean and standard deviation of the
error on the test set. We tried two levels of random noise $q=0.1$ and
$q=0.2\%$.  The boosting algorithms are run for (at most) 500
iterations.  For Robustboost we use: $\theta=1.0$, $\sigma=0.1$, for
$q=0.1$ $\epsilon=0.15$, for $q=0.2$ $\epsilon=0.25$. For these
settings of $\epsilon$ Robustboost terminates after 100-300
iterations.

\begin{tabular}{|l|c|c|c|c|}
\hline
base & q  & Ada & Logit &  Robust \\
\hline
stump &$10$& $19.3 \pm 1.0$ & $15.9 \pm 0.9$ & $13.5 \pm 0.8$ \\
\hline
2tree &$10$& $21.4 \pm 1.2$ & $18.7 \pm 1.1$ & $14.8 \pm 1.0$ \\
\hline
\hline
stump &$20$& $29.4 \pm 1.2$ & $26.7 \pm 1.3$  & $23.8 \pm 1.1$ \\
\hline
2Tree &$20$& $32.3 \pm 1.0$ & $29.3 \pm 0.8$  & $25.3 \pm 0.9$ \\
\hline
\end{tabular}

As in the previous section Robustboost performs significantly better
than Logitboost which is better than Adaboost. The relative
performance of the algorithms is consistent with this table in all 15
repeats of the experiment. Using 2 level decision trees is
consistently worse than using stumps. While significant, the
difference between Robustboost and Logitboost is smaller than it was
in the previous section, we conjecture that this is because the
distribution here is much more symmetric which decreases the biasing
effect of the examples with large negative margins.

Continuing only with stumps, we report the error of the generated
classifiers relative to the uncorrupted labels. The relative
performance here is the same, with Robustboost leading the way.

\begin{tabular}{|l|c|c|c|c|}
\hline
q  & Ada & Logit &  Robust \\
\hline
$10$& $11.5 \pm 1.1$ & $7.1 \pm 0.7$ & $4.3 \pm 0.4$ \\
\hline
$20$& $15.6 \pm 1.2$ & $11.2 \pm 1.0$  & $6.5 \pm 1.0$ \\
\hline
\end{tabular}

A potentially more important aspect of the classifier generated by
Robustboost is that it's predictions that are given with large margins
are very trustworthy. In the following table we report the fraction of
the test set on which the absolute value of the score is smaller than
$\theta$ (low margin examples) and the error rate on the remaining
examples relative to the uncorrupted test data.

\begin{tabular}{|l|c|c|c|c|}
\hline
q  & low margin & clean err \\
\hline
$10$& $10.5 \pm 0.6$ & $0.9 \pm 0.2$ \\
\hline
$20$& $10.2 \pm 0.7$ & $2.4 \pm 0.6$ \\
\hline
\end{tabular}

Once more, we see that Robustboost is capable of detecting most of the
incorrect labels for the examples with large margins. 

\section{conclusions}
\label{sec:conclusion}
We present evidence that Robustboost is more robust against label
noise than either Logitboost or Adaboost. More experiments using
synthetic and real-world datasets are needed to verify this claim.

The effectiveness of Robustboost suggest that after an approximate
classifier has been learned it can be beneficial to down-weight
examples that are far from the decision boundary regardless of their
label. This suggests new directions for active learning which we are
currently investigating.

\iffalse
\begin{figure}[t]
%\vskip 0.2in
\begin{center}
\centerline{\includegraphics[width=\columnwidth]{Figures/Stumps0_1.pdf}}
\caption{}
\label{fig:Mease0.1comparison}
\end{center}
\vskip -0.4in
\end{figure} 
\fi

\begin{small}
\bibliography{bib}
\bibliographystyle{mlapa}
\end{small}

\end{document}